\documentclass[12pt]{article}

\usepackage{sbc-template}
\usepackage{graphicx,url}
\usepackage[utf8]{inputenc}
\usepackage{multirow}
\usepackage{booktabs} 
\usepackage{subcaption}

\newcommand{\orcidID}[1]{\textnormal{\normalsize\textsuperscript{\small\texttt{#1}}}}
 
\title{BLUEX Revisited: Enhancing Benchmark Coverage with Automatic Captioning}

\author{João Guilherme Alves Santos\inst{1}\orcidID{0000-0001-5307-5338} \and
Giovana Kerche Bonás\inst{1,2}\\\orcidID{0009-0001-9460-8353} \and
Thales Sales Almeida\inst{1,2}\orcidID{0009-0006-9568-9331}}

\address{Instituto de Computação (IC) -- Universidade Estadual de Campinas (UNICAMP)\\
 Campinas -- SP -- Brazil
\nextinstitute
  Maritaca AI\\
  maritaca.ai
  \email{j199624@dac.unicamp.br, g216832@dac.unicamp.br,
  t224732@dac.unicamp.br}
}

\begin{document} 

\maketitle

\begin{abstract}
With the growing capabilities of Large Language Models (LLMs), there is an increasing need for robust evaluation methods, especially in multilingual and non-English contexts. We present an updated version of the BLUEX dataset, now including 2024-2025 exams and automatically generated image captions using state-of-the-art models, enhancing its relevance for data contamination studies in LLM pretraining. Captioning strategies increase accessibility to text-only models by more than 40\%, producing 1,422 usable questions, more than doubling the number in the original BLUEX. We evaluated commercial and open-source LLMs and their ability to leverage visual context through captions.

\end{abstract}

\section{Introduction} \label{sec:introduction}

Large Language Models (LLMs) have made remarkable progress in recent years, demonstrating impressive capabilities across a wide range of natural language processing tasks, such as code generation and assistance~\cite{nam2024using,liu2023your,chen2021evaluating,zhang2023planning}, question answering~\cite{singhal2025toward,petroni2019language,lazaridou2022internet,almeida2025tiebe}, open-ended conversation~\cite{openai2024gpt4technicalreport,grattafiori2024llama3,abonizio2024sabia,ouyang2022training}, and summarization~\cite{zhang2024benchmarking,zhang2024comprehensive,chang2023booookscore}.
As these models become increasingly capable of performing complex reasoning and generating coherent, context-aware responses, robust benchmarks play a central role in assessing their true capabilities. Beyond measuring surface-level accuracy, well-designed benchmarks can help uncover how models handle ambiguity, reason through multiple steps, and generalize across diverse topics and linguistic styles. 

Considering real-world utility, one domain that naturally demands advanced understanding and complex reasoning is standardized education. High-stakes exams often require students to interpret complex textual information --sometimes alongside diagrams, graphs, illustrations, or images-- and respond to questions that require factual knowledge and inferential thinking. These settings provide rich and authentic challenges where textual understanding and reasoning across domains directly impact task success. Because such assessments are carefully designed to evaluate specific cognitive skills and knowledge, they offer a grounded, purpose-driven context to evaluate the capabilities of large language models.

In this work, we introduce an expanded and updated version of BLUEX~\cite{almeida2023bluex}, a benchmark comprising over 1,000 multiple-choice questions from the entrance exams of Brazil’s top universities and the top 500 worldwide~\cite{times2024,shanghai2024}, Unicamp and USP, administered between 2018 and 2023. Each question includes text, answer options, and associated images. Our main objective is to evaluate how large language models perform in multimodal educational tasks, and to investigate the impact of different image captioning strategies on their performance. To do this, we expand the original dataset both quantitatively and temporally, including two additional years (2024 and 2025) and generate image captions for all visual elements using GPT4o.


We generate image descriptions using GPT-4o under two distinct conditions: \textit{Blind captions}, where captions are generated solely based on visual content without access to accompanying text; and \textit{Context captions}, where captions are generated with access to the associated question and alternatives, allowing context-aware interpretations. This setup not only broadens the dataset with image-associated questions that were previously unanswerable by non-multimodal LLMs, but also provides a controlled experimental setting to analyze how context-aware image captioning affects LLM performance, under the hypothesis that contextualized captions --though often shorter-- may lead to equal or even superior model accuracy by focusing on task-relevant visual elements.

Our findings show that many current models now reach scores high enough to surpass the admission thresholds of over 90\% of undergraduate programs at Unicamp and USP, highlighting the rapid progress of LLMs in handling complex reasoning tasks. By releasing this updated dataset\cite{bluex_dataset}, along with the evaluation code\cite{bluex_code}, we aim to provide a more realistic and multimodal benchmark to evaluate LLMs in educational contexts, while offering empirical insights into how different captioning strategies influence their ability to interpret and reason about image-based content. Additionally, by converting image-based questions into caption-based ones, our benchmark extends its applicability to non-multimodal models, enabling broader participation in multimodal tasks and allowing researchers to isolate the effects of visual grounding through controlled captioning strategies.

\section{Related work} \label{sec:relatedwork}

\subsection{LLM standard test evaluation}

A growing body of work has evaluated large language models (LLMs) using standardized exams as proxies for human-level reasoning and real-world task performance. Such as example, MMLU (Massive Multitask Language Understanding) \cite{hendrycks2021measuringmassivemultitasklanguage} contains 57 tasks --that include STEM questions, humanities, social sciences and other fields-- and represents a standard reference for measuring academic proficiency in LLMs, evaluating factual knowledge, reasoning and problem-solving abilities enabling evaluate models through both zero-shot and few-shot settings.

Moreover, AGIEval\cite{zhong2023agieval} also presents a similar landscape, evaluating LLMs in real-world cognitive challenges through standardized exams, such as college entrance exams, math competitions, and lawyer qualification tests. Furthermore, GPQA\cite{rein2024gpqa} raises the bar for evaluating LLMs by introducing a highly challenging dataset of 448 multiple-choice questions authored by specialists from a range of academic domains. These questions are designed to be extremely difficult --even for experts-- and resist simple retrieval strategies, making GPQA a valuable benchmark for assessing deep reasoning and scalable oversight in advanced AI systems.

Delving into Portuguese-language evaluations, a few large-scale benchmarks exist to assess LLMs in Portuguese. But among them, BLUEX~\cite{almeida2023bluex} stands out as a comprehensive and challenging dataset focused on Brazilian university entrance exams. Comprising more than 1,000 multiple-choice questions from Brazil's Unicamp and USP entrance exams (2018–2023), it is designed to reflect the complexity of real-world educational assessments, including a significant portion (~40\%) of questions with visual components. This benchmark serves as the foundation for our present work, which extends BLUEX by adding two more years of exam data and incorporating a captioning pipeline, enabling a richer evaluation, particularly for models without native vision capabilities.

Other important efforts include benchmarks based on structured assessments traditionally used in Brazil, such as the Brazilian Bar Examination (OAB)~\cite{delfino2017passing} and the National High School Exam (ENEM)\cite{silveira2017university}, both of which have been used to explore reasoning and comprehension of language models.


\subsection{Multimodal Benchmarks}

Recent advances in multimodal large language models (MLLMs) have highlighted the need for specialized benchmarks that can rigorously assess their visual and reasoning abilities. To meet this demand, recent studies have introduced diagnostic evaluations that go beyond raw performance, aiming to uncover specific strengths and weaknesses of models in complex multimodal settings. Benchmarks such as SEED-Bench~\cite{li2023seed} and its successor studies\cite{li2024seed2} have focused on systematically evaluating the multimodal reasoning capabilities of large language models. The benchmark consists of 24,000 multiple-choice questions covering 27 evaluation dimensions, including topics such as charts, visual mathematics, and free-form interleaved image-text reasoning. By restricting outputs to A/B/C/D choices and combining human annotation with automated filtering, it provides a scalable and objective evaluation framework, revealing that even top models achieve only around 60\% accuracy and highlighting major challenges in complex visual and reasoning tasks.

Similarly, the Perception Test~\cite{NEURIPS2023_8540fba4} extends multimodal evaluation into the video domain, proposing a diagnostic benchmark that measures fine-grained perceptual abilities and temporal reasoning across thousands of annotated videos. Instead of static images, models are challenged with dynamic visual scenes requiring continuous understanding, spatial awareness, and causal inference. In parallel, Importantly, MM-SafetyBench~\cite{liu2024mmsafety} reveals an even deeper layer of vulnerability, showing that MLLMs can be easily compromised through query-relevant images, even when their underlying LLMs are safety-aligned. This demonstrates that multimodal safety demands specific, dedicated attention, beyond what is currently done for text-only models. 



Building upon these insights, this expanded dataset provides a valuable resource for future studies aiming to assess linguistic models in Portuguese and in authentic educational contexts. In this way, our contribution supports the development of benchmarks that move beyond English-centric paradigms, enriching the tools available for cross-linguistic and culturally grounded multimodal evaluation.

\subsection{Caption generation by Multimodal Models}


As a strategy to navigate towards vision and language modalities, recent works have explored replacing or augmenting images with textual descriptions. A notable example, \textit{"Evaluating GPT-4’s Vision Capabilities on Brazilian University Admission Exams"}\cite{pires2023evaluatinggpt4svisioncapabilities}, focuses on Brazil's national high school exam (ENEM), evaluating GPT-4 in recent ENEM editions, incorporating both textual and visual elements. A key finding is that text captions transcribing visual content often outperform the direct use of images. This highlights the potential of captioning as a powerful bridge between visual data and language models, suggesting that carefully crafted textual descriptions can unlock complex reasoning in models without requiring full visual processing. Building on this insight, our work integrates a captioning pipeline into the BLUEX, enabling a more targeted and scalable evaluation of LLMs on visually grounded academic tasks. 

Some of these efforts can also be represented by projects such as Image Textualization (IT)~\cite{pi2024image}, a framework that collaborates with multiple expert vision models with MLLMs to automatically generate high-quality and detailed image descriptions. In a similar vein, Bianco et al. \cite{bianco2023improving} propose a method to enhance the quality of image captions by integrating the output of various state-of-the-art models using LLMs, providing richer captions. In the same direction, TIFA~\cite{hu2023tifa} also explores that field by making an evaluation that demonstrates that high-quality captions can significantly impact model performance in visual tasks.

\section{Methodology}
\begin{figure}[h]
    \centering
    \includegraphics[width=1\columnwidth]{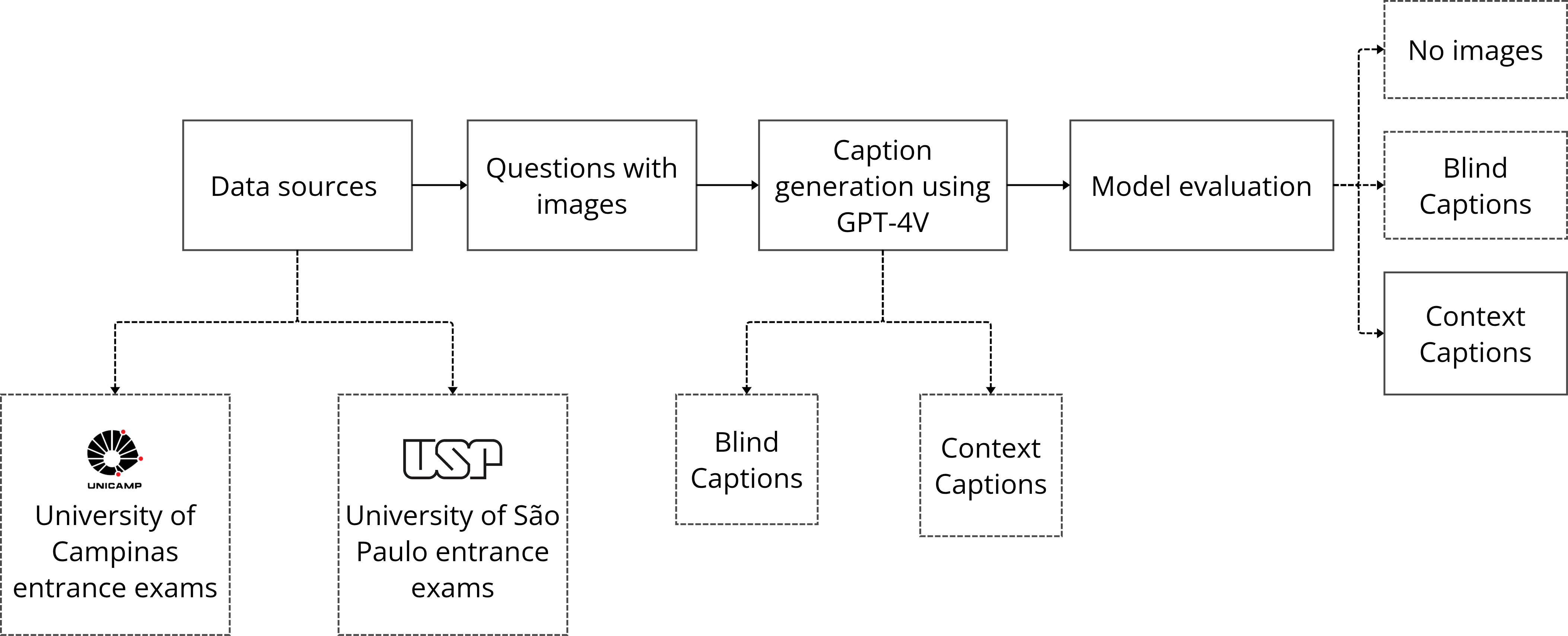}
    \caption{Overview of the benchmark construction pipeline.}
    \label{fig:diagram}
\end{figure}

This section describes the pipeline to create the benchmark, which is illustrated in Figure \ref{fig:diagram}.

The updated dataset is publicly available~\cite{bluex_dataset} --as well as the evaluation code~\cite{bluex_code}-- and includes the original visual content, two types of generated captions, full question text, answer choices, correct answer, and metadata relevant to the nature of the assessment for each question. All data were manually curated and classified. This structure enables both rigorous evaluation and reproducible experimentation with multimodal and text-only models.

\subsection{Dataset Selection}
We use the entire Brazilian Leading Universities Entrance eXams (BLUEX) collection introduced by \cite{almeida2023bluex}, designed to address the scarcity of high-quality datasets in Portuguese by compiling entrance exam questions from Brazil's two most competitive universities. Additionally, we expanded the dataset to incorporate the most recent exams from 2024 and 2025 and provided new generated captions for visual content in Portuguese.

\subsection{Caption Generation}
Roughly 43\% of the original BLUEX questions included images and were therefore inaccessible to text-only LLMs. To make these items evaluable, we generate Portuguese textual descriptions for every image with GPT-4o under two settings ~\cite{hurst2024gpt4o}:
\begin{enumerate}
    \item Blind Captions: GPT-4o generates captions based solely on the image, without any contextual information from the associated question.
    \item Context Captions: GPT-4o is provided with both the image and its associated question, enabling it to generate context-aware descriptions.
\end{enumerate}

\begin{figure}[h]
  \centering
  \begin{subfigure}[b]{0.50\linewidth}
    \centering
    \includegraphics[width=\linewidth]{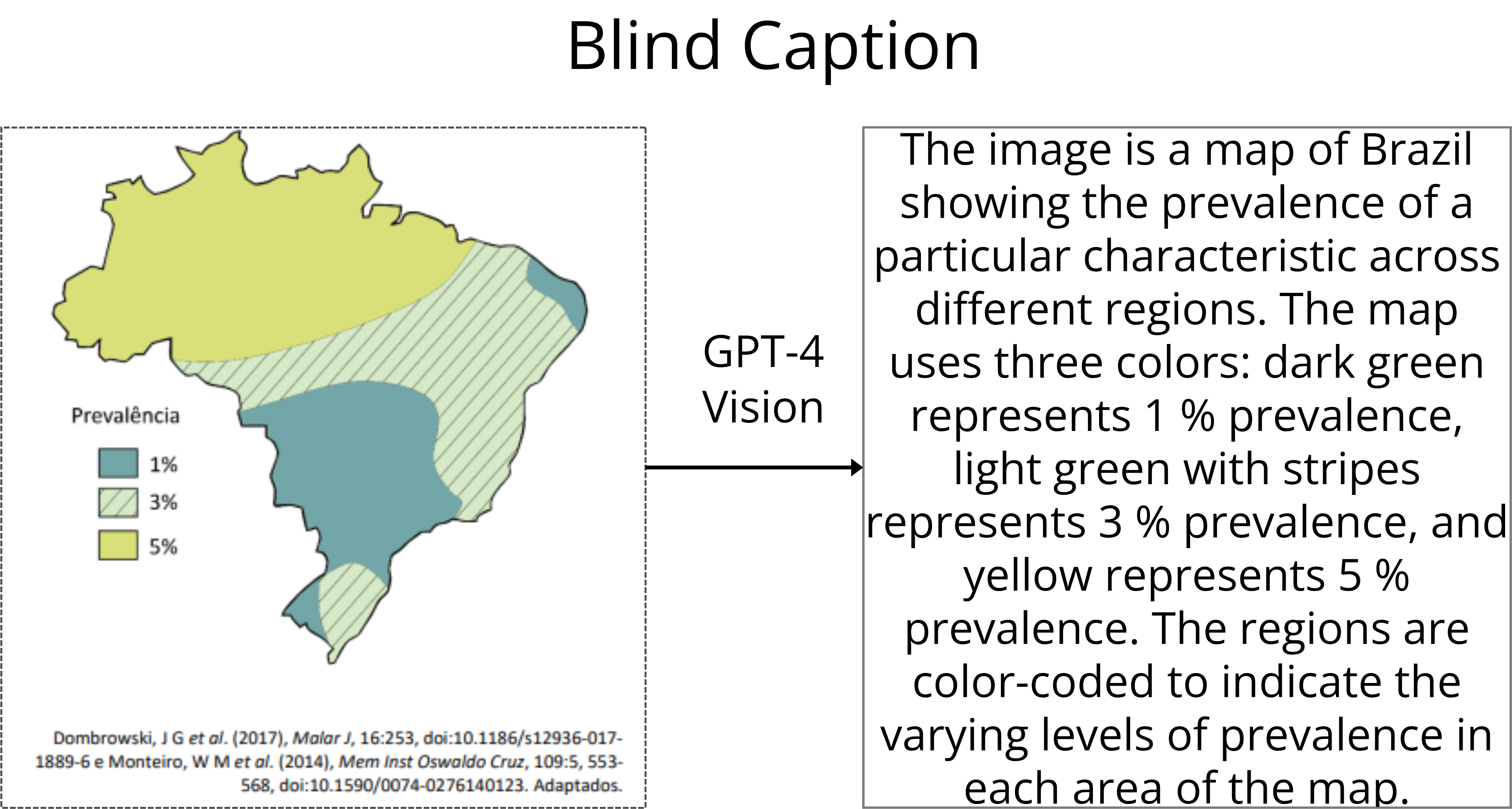}
    \subcaption{Blind caption}\label{fig:blind}
  \end{subfigure}
  \hfill                               
  \begin{subfigure}[b]{0.49\linewidth}
    \centering
    \includegraphics[width=\linewidth]{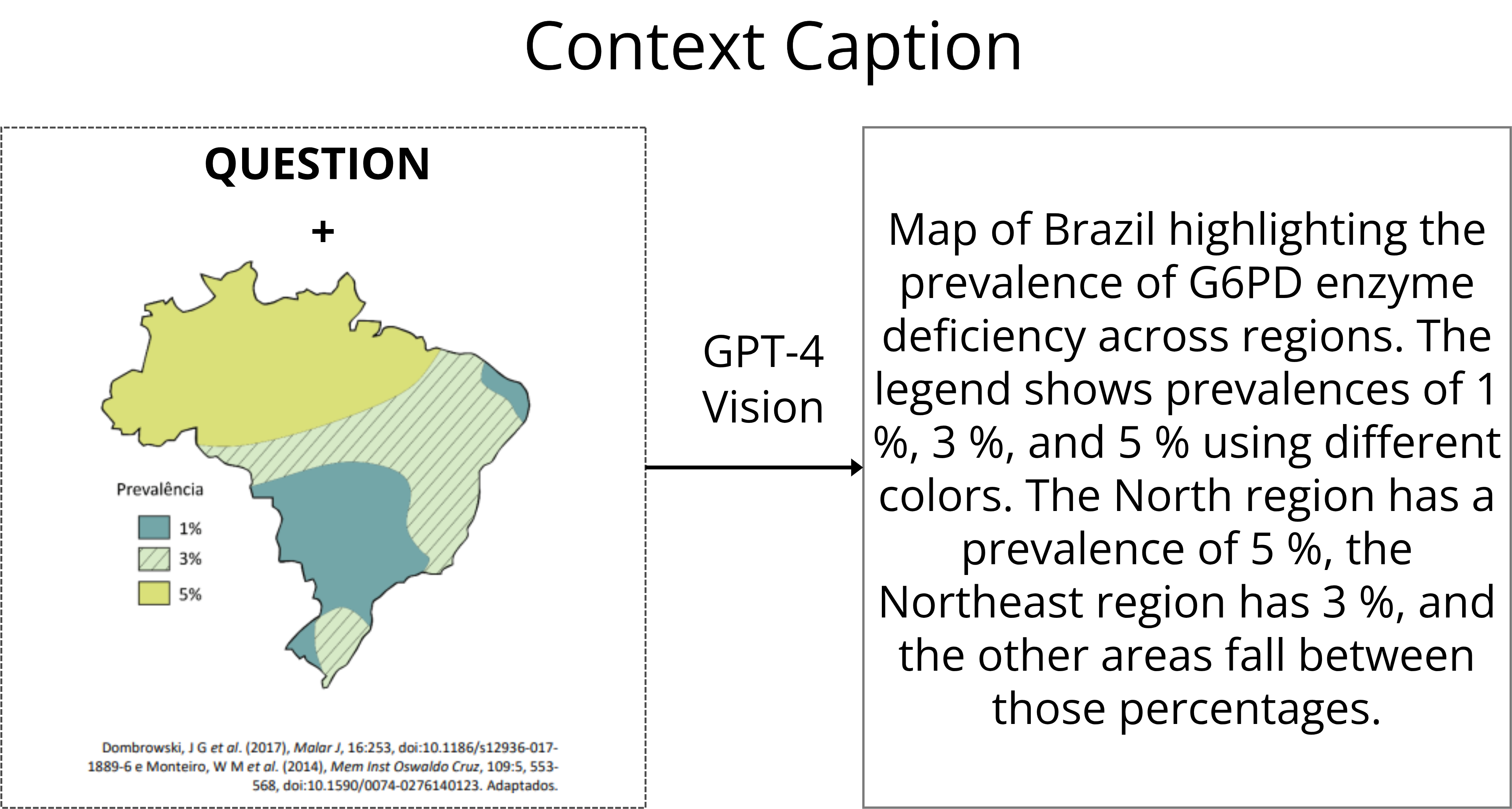}
    \subcaption{Context caption}\label{fig:context}
  \end{subfigure}
    \caption{Comparative example of blind and context captioning. \textit{Captions were generated originally in Portuguese; presented in English for convenience.}}

  \label{fig:example_blind_x_context_captions}
\end{figure}

Figure \ref{fig:example_blind_x_context_captions} illustrates a comparative example of the two captioning strategies, exposing that context captions such as shown in Figure \ref{fig:context} tended to be significantly shorter than blind captions in Figure \ref{fig:blind}. This phenomenon likely arises because GPT-4o, when given access to the question, focuses on the visual elements most relevant to answering the task.

\subsection{Model Evaluation}
To assess how effectively language models leverage the captions when answering exam questions, we conducted evaluations across three distinct experimental conditions. These conditions vary the amount and type of information provided to the models. Specifically, we evaluated models using the following configurations:
\begin{enumerate}
    \item No images: The model receives only the textual content of the question, without any accompanying visual information (neither images nor captions). This baseline measures performance in purely textual scenarios.

    \item Blind Captions: In place of actual images, the model is provided with the corresponding blind captions --descriptive captions generated without question context.

    \item Context Captions: Here, the models receive context-aware captions --captions generated by GPT-4o using both the question and image as inputs. 

\end{enumerate}
Through these experiments, we aim to quantify how visual context and caption specificity affect model performance, thereby offering clearer insight into each model’s effectiveness.

\subsection{Caption and Image Statistics}  
\begin{figure}[h]
    \centering
    \includegraphics[width=0.7\columnwidth]{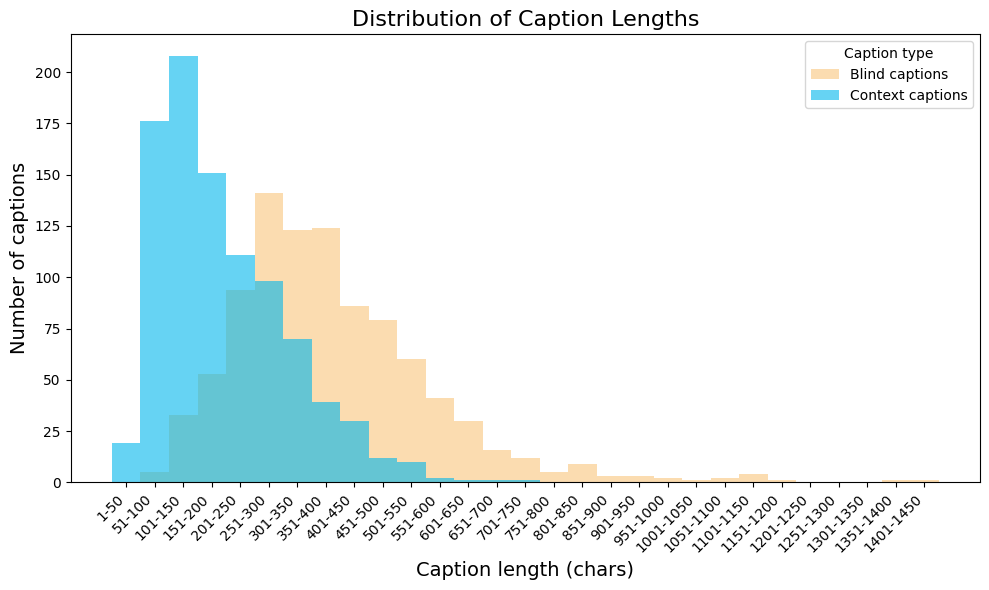}
    \caption{Distribution of caption lengths for blind and context captions.}
    \label{fig:caption_lengths}
\end{figure}

Figure~\ref{fig:caption_lengths} presents the distribution of the caption lengths produced by GPT‑4V under the two prompting strategies. As expected, context captions are markedly shorter than blind captions, because the question context guides the model to mention only the information most relevant for solving the problem. 

Table \ref{tab:image_prevalence} organizes all BLUEX questions into the four ENEM macro-areas: Natural Sciences (Biology, Chemistry, Physics), Human Sciences (History, Geography, Philosophy, Sociology), Languages (Portuguese and English), and Mathematics - and reports, for each year from 2018 to 2025, the number of items that do and do not include associated images. The proportion of image-based questions is consistently high across the entire period, underscoring the value of the captioning step. Once textual descriptions are available, every image question becomes accessible to text-only models. Per-area subtotals do not equal the grand total because a single question can be annotated with more than one subject, reflecting its inherently interdisciplinary character.

\begin{table}[h]
\centering
\caption{Number of BLUEX questions by ENEM macro area and image distribution}
\label{tab:image_prevalence}
\resizebox{\textwidth}{!}{
\begin{tabular}{lcccc|cc|c}
\hline
\multicolumn{1}{c}{} &
  \multicolumn{4}{c|}{\textbf{Subject Distribution}} &
  \multicolumn{2}{l|}{\textbf{Image distribution}} &
  \multicolumn{1}{l}{} \\ \hline
\multicolumn{1}{l|}{\textbf{Year}} &
  \textbf{\begin{tabular}[c]{@{}c@{}}Natural \\ Sciences\end{tabular}} &
  \textbf{\begin{tabular}[c]{@{}c@{}}Human \\ Sciences\end{tabular}} &
  \textbf{Languages} &
  \textbf{Mathematics} &
  \begin{tabular}[c]{@{}c@{}}\textbf{With}\\ \textbf{Images}\end{tabular} &
  \begin{tabular}[c]{@{}c@{}}\textbf{Without}\\ \textbf{Images}\end{tabular} &
  \textbf{Total} \\ \hline
\multicolumn{1}{l|}{2018}           & 72  & 49  & 46  & 25  & 67 & 113 & 180  \\
\multicolumn{1}{l|}{2019}           & 68  & 53  & 50  & 31  & 90 & 90 & 180  \\
\multicolumn{1}{l|}{2020}           & 69  & 51  & 50  & 33  & 69 & 111 & 180  \\
\multicolumn{1}{l|}{2021}           & 82  & 64  & 64  & 40  & 85 & 149 & 234  \\
\multicolumn{1}{l|}{2022}           & 51  & 48  & 43  & 26  & 74 & 88 & 162  \\
\multicolumn{1}{l|}{2023}           & 59  & 50  & 41  & 24  & 75 & 87 & 162  \\
\multicolumn{1}{l|}{2024}           & 52  & 57  & 42  & 26  & 76 & 86 & 162  \\
\multicolumn{1}{l|}{2025}           & 53  & 62  & 54  & 25  & 74 & 88 & 162  \\ \hline
\multicolumn{1}{l|}{\textbf{Total}} & 506 & 434 & 390 & 230 & 610 & 812 & 1422 \\ \hline
\end{tabular}
}
\end{table}

\subsection{Implementation Details}

For each question, the input prompt consists of the full textual content (question statement and multiple-choice alternatives) --formatted as it originally appeared in the exam. When image captions are included in the evaluation, they are inserted into the prompt in the same position as the original image in the source exam layout.
We use lm-evaluation-harness~\cite{eval-harness} to evaluate open-source models, using FP16 precision. Small models (up to 14B parameters) were executed on a machine with 2× A6000 GPUs. Llama 3.3 (70B) and Qwen 2.5 (72B) were executed on a machine with 2× A100 80GB GPUs. The Deepseek-v3 and Llama 4 models were executed via the Together AI API~\footnote{\url{https://www.together.ai/}} due to their high parameter count.

\section{Results}

We evaluated several language models, separating them into two categories, small (every model with 14B parameters or less) and big models (70B and more parameters), as well as commercial models in sabiazinho-3, sabia-3~\cite{abonizio2024sabia} from Maritaca AI, and GPT-4o and GPT-4o-mini~\cite{hurst2024gpt4o} from openAI. Our complete results are presented in Table~\ref{tab:full_results}, which reports the results for three partitions of the dataset: questions without images, questions with images, and considering all questions in the benchmark. For partitions that involve images, we test using both context-aware and blind captions, as well as providing no caption at all.


\begin{table}[h]
\centering
\caption{Performance of all the tested models in the BLUEX benchmark}
\label{tab:full_results}
\resizebox{1\textwidth}{!}{
\begin{tabular}{@{}lccccccc@{}}
\toprule
\multicolumn{1}{l|}{\multirow{2}{*}{Model}} &
  \multicolumn{1}{c|}{\begin{tabular}[c]{@{}c@{}}Questions without \\ images\end{tabular}} &
  \multicolumn{3}{c|}{\begin{tabular}[c]{@{}c@{}}Questions with \\ images\end{tabular}} &
  \multicolumn{3}{c}{All Questions} \\ \cmidrule(l){2-8} 
\multicolumn{1}{l|}{} &
  \multicolumn{1}{c|}{\textbf{\begin{tabular}[c]{@{}c@{}}No \\ Caption\end{tabular}}} &
  \textbf{\begin{tabular}[c]{@{}c@{}}Context \\ Captions\end{tabular}} &
  \textbf{\begin{tabular}[c]{@{}c@{}}Blind \\ Captions\end{tabular}} &
  \multicolumn{1}{c|}{\textbf{\begin{tabular}[c]{@{}c@{}}No \\ Caption\end{tabular}}} &
  \textbf{\begin{tabular}[c]{@{}c@{}}Context \\ Captions\end{tabular}} &
  \textbf{\begin{tabular}[c]{@{}c@{}}Blind \\ Captions\end{tabular}} &
  \textbf{\begin{tabular}[c]{@{}c@{}}No \\ Caption\end{tabular}} \\ \midrule
\multicolumn{8}{l}{\textbf{Comercial models}} \\ 

\multicolumn{1}{l|}{Sabia-3~\cite{abonizio2024sabia}} &
  \multicolumn{1}{c|}{\textbf{0.852}} &
  0.701 &
  0.695 &
  \multicolumn{1}{c|}{0.616} &
  \textbf{0.787} &
  \textbf{0.784} &
  0.750 \\
\multicolumn{1}{l|}{GPT-4o~\cite{hurst2024gpt4o}} &
  \multicolumn{1}{c|}{0.807} &
  \textbf{0.718} &
  \textbf{0.729} &
  \multicolumn{1}{c|}{\textbf{0.683}} &
  0.769 &
  0.774 &
  \textbf{0.754} \\
\multicolumn{1}{l|}{GPT-4o-mini~\cite{hurst2024gpt4o}} &
  \multicolumn{1}{c|}{0.785} &
  0.642 &
  0.627 &
  \multicolumn{1}{c|}{0.589} &
  0.724 &
  0.717 &
  0.701 \\
  \multicolumn{1}{l|}{Sabiazinho-3~\cite{abonizio2024sabia}} &
  \multicolumn{1}{c|}{0.756} &
  0.642 &
  0.652 &
  \multicolumn{1}{c|}{0.627} &
  0.707 &
  0.712 &
  0.701 \\
  \midrule
\multicolumn{8}{l}{\textbf{Large Open source models}} \\ 
\multicolumn{1}{l|}{DeepSeek-V3~\cite{liu2024deepseek3}} &
  \multicolumn{1}{c|}{\textbf{0.841}} &
  \textbf{0.739} &
  \textbf{0.741} &
  \multicolumn{1}{c|}{\textbf{0.668}} &
  \textbf{0.797} &
  \textbf{0.798} &
  \textbf{0.767} \\
\multicolumn{1}{l|}{Llama-4-Scout} &
  \multicolumn{1}{c|}{0.758} &
  0.644 &
  0.665 &
  \multicolumn{1}{c|}{0.589} &
  0.709 &
  0.718 &
  0.685 \\
\multicolumn{1}{l|}{Llama-4-Maverick} &
  \multicolumn{1}{c|}{0.820} &
  0.731 &
  0.731 &
  \multicolumn{1}{c|}{0.658} &
  0.781 &
  0.781 &
  0.750 \\
\multicolumn{1}{l|}{Llama-3.3-70B-Instruct~\cite{dubey2024llama3}} &
  \multicolumn{1}{c|}{0.769} &
  0.647 &
  0.657 &
  \multicolumn{1}{c|}{0.603} &
  0.717 &
  0.721 &
  0.697 \\
\multicolumn{1}{l|}{Qwen2.5-72B-Instruct~\cite{yang2024qwen2}} &
  \multicolumn{1}{c|}{0.796} &
  0.695 &
  0.693 &
  \multicolumn{1}{c|}{0.650} &
  0.752 &
  0.752 &
  0.733 \\ \midrule
\multicolumn{8}{l}{\textbf{Small open source models}} \\ 
\multicolumn{1}{l|}{Qwen 2.5-14B~\cite{yang2024qwen2}} &
  \multicolumn{1}{c|}{\textbf{0.745}} &
  \textbf{0.637} &
  \textbf{0.640} &
  \multicolumn{1}{c|}{\textbf{0.614}} &
  \textbf{0.699} &
  \textbf{0.700} &
  \textbf{0.689} \\
\multicolumn{1}{l|}{Qwen 2.5-7B~\cite{yang2024qwen2}} &
  \multicolumn{1}{c|}{0.676} &
  0.568 &
  0.558 &
  \multicolumn{1}{c|}{0.547} &
  0.630 &
  0.626 &
  0.621 \\
\multicolumn{1}{l|}{Qwen 2.5-3B~\cite{yang2024qwen2}} &
  \multicolumn{1}{c|}{0.587} &
  0.530 &
  0.507 &
  \multicolumn{1}{c|}{0.483} &
  0.563 &
  0.553 &
  0.542 \\
\multicolumn{1}{l|}{Qwen 2.5-1.5B~\cite{yang2024qwen2}} &
  \multicolumn{1}{c|}{0.524} &
  0.427 &
  0.409 &
  \multicolumn{1}{c|}{0.384} &
  0.482 &
  0.475 &
  0.464 \\
\multicolumn{1}{l|}{Falcon-10B~\cite{Falcon3}} &
  \multicolumn{1}{c|}{0.669} &
  0.575 &
  0.560 &
  \multicolumn{1}{c|}{0.527} &
  0.628 &
  0.622 &
  0.608 \\
\multicolumn{1}{l|}{Falcon-7B~\cite{Falcon3}} &
  \multicolumn{1}{c|}{0.612} &
  0.534 &
  0.524 &
  \multicolumn{1}{c|}{0.509} &
  0.578 &
  0.574 &
  0.568 \\
\multicolumn{1}{l|}{Falcon-3B~\cite{Falcon3}} &
  \multicolumn{1}{c|}{0.455} &
  0.422 &
  0.392 &
  \multicolumn{1}{c|}{0.389} &
  0.441 &
  0.428 &
  0.427 \\
\multicolumn{1}{l|}{Falcon-1B~\cite{Falcon3}} &
  \multicolumn{1}{c|}{0.266} &
  0.266 &
  0.253 &
  \multicolumn{1}{c|}{0.240} &
  0.266 &
  0.260 &
  0.255 \\
\multicolumn{1}{l|}{Llama-3.1-8B~\cite{dubey2024llama3}} &
  \multicolumn{1}{c|}{0.595} &
  0.473 &
  0.466 &
  \multicolumn{1}{c|}{0.448} &
  0.542 &
  0.539 &
  0.532 \\
 \bottomrule
\end{tabular}
}
\end{table}

\begin{figure}
    \centering
    \includegraphics[width=0.6\linewidth]{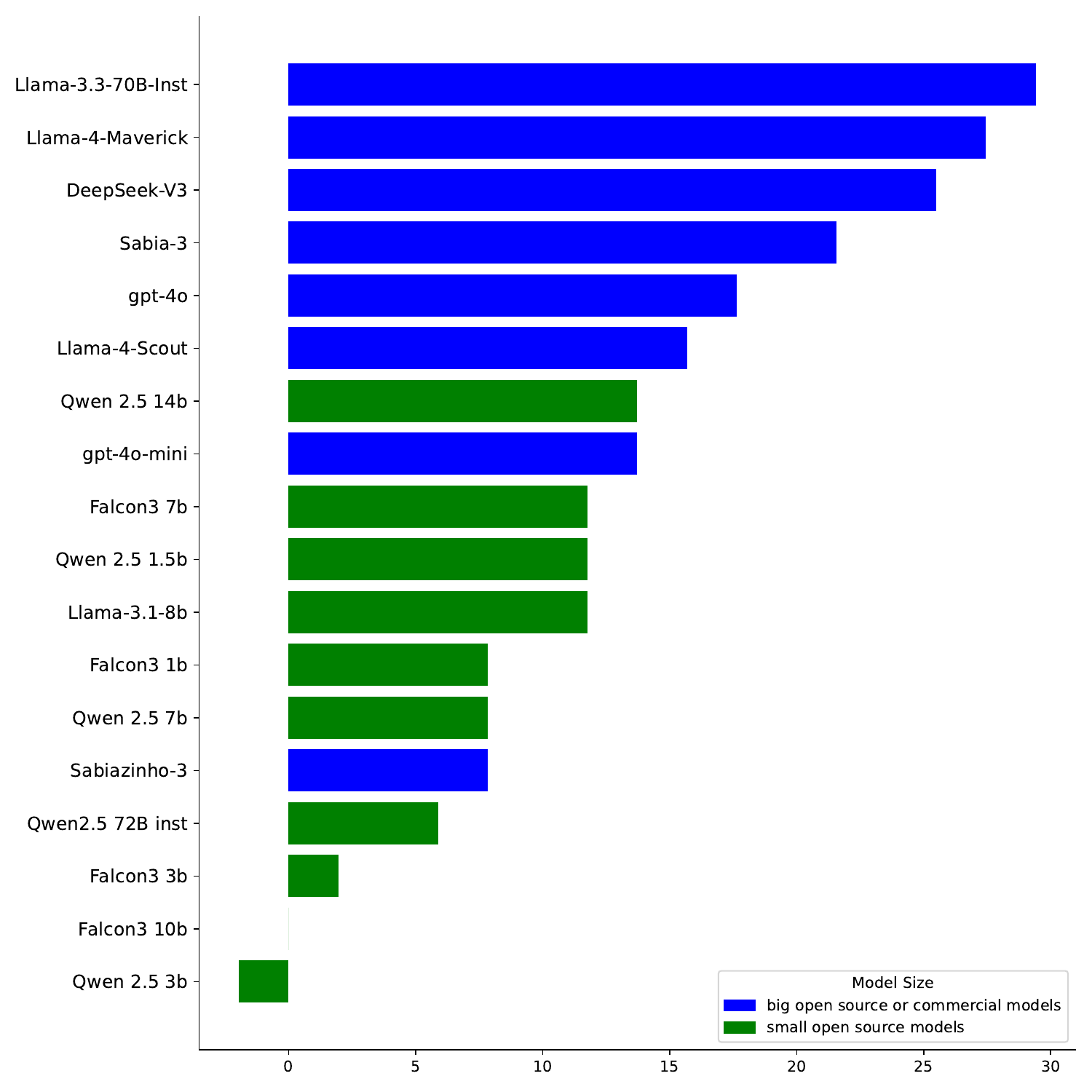}
    \caption{Accuracy gain in BLUEX questions that have images as alternatives. The graph shows the accuracy gain for each model by providing context captions for the images, compared to the blind performance.}
    \label{fig:bar_performance_increase_caption}
\end{figure}

%

Sabia-3 achieves the best performance among commercial models on questions without images, surpassing GPT-4o, the second-best commercial model, by 5 points. However, this performance lead does not hold when looking at questions with images, where GPT-4o shows the best performance. Sabia-3 and GPT-4o perform similarly across all questions, with Sabia-3 leading by a small margin.

Within large open-source models, DeepSeek-V3 is the most robust model across all categories, presenting the highest accuracy in questions without images and performing consistently well across caption scenarios in questions with images. LLama-4-maverick ranks as the second-best performance. Note that both are mixture of experts(MOE) models with more than 400B total parameters. Qwen-2.5-72B~\cite{yang2024qwen2} ranks as the third best model among large open source models, a very competitive performance for a model given it has less than 25\% total parameters compared to the first two models.

Among small open-source models, Qwen 2.5-14B~\cite{yang2024qwen2} stands out significantly, showcasing strong capabilities in all scenarios. This model notably surpasses other smaller models, such as Falcon and Llama-3. Generally, the Qwen 2.5 family of models performs better than their counterparts of the same size.

In general, we observe a slight variation between the performance of models with context or blind captions, even with the context captions being on average half the size, this indicates that the additional context provided when creating context captions did not necessarily make the caption more informative, but rather allowed it to be more concise and focused only on the relevant aspects of the image. Meanwhile, the blind description of the image used in the blind captions does seem to provide enough information to answer some questions, at the expense of longer captions.

\subsection{How Effectively Models Use Captions}

To better understand the impact of providing captions to models, we specifically analyzed questions whose alternatives consisted solely of images --questions that would be impossible to answer correctly without additional visual context. For these cases, we evaluated how much each model's performance improved when provided with context-aware captions. 

Figure~\ref{fig:bar_performance_increase_caption} illustrates these improvements. Upon receiving captions, most models exhibited a performance gain of at least 10 accuracy points. Additionally, we observed that larger models tended to benefit more significantly from captions, likely because interpreting and reasoning from captions require higher cognitive capabilities that scale with model size.

\subsection{Comparing Models with Human Performance}

\begin{figure}
    \centering
    \includegraphics[width=0.995\linewidth]{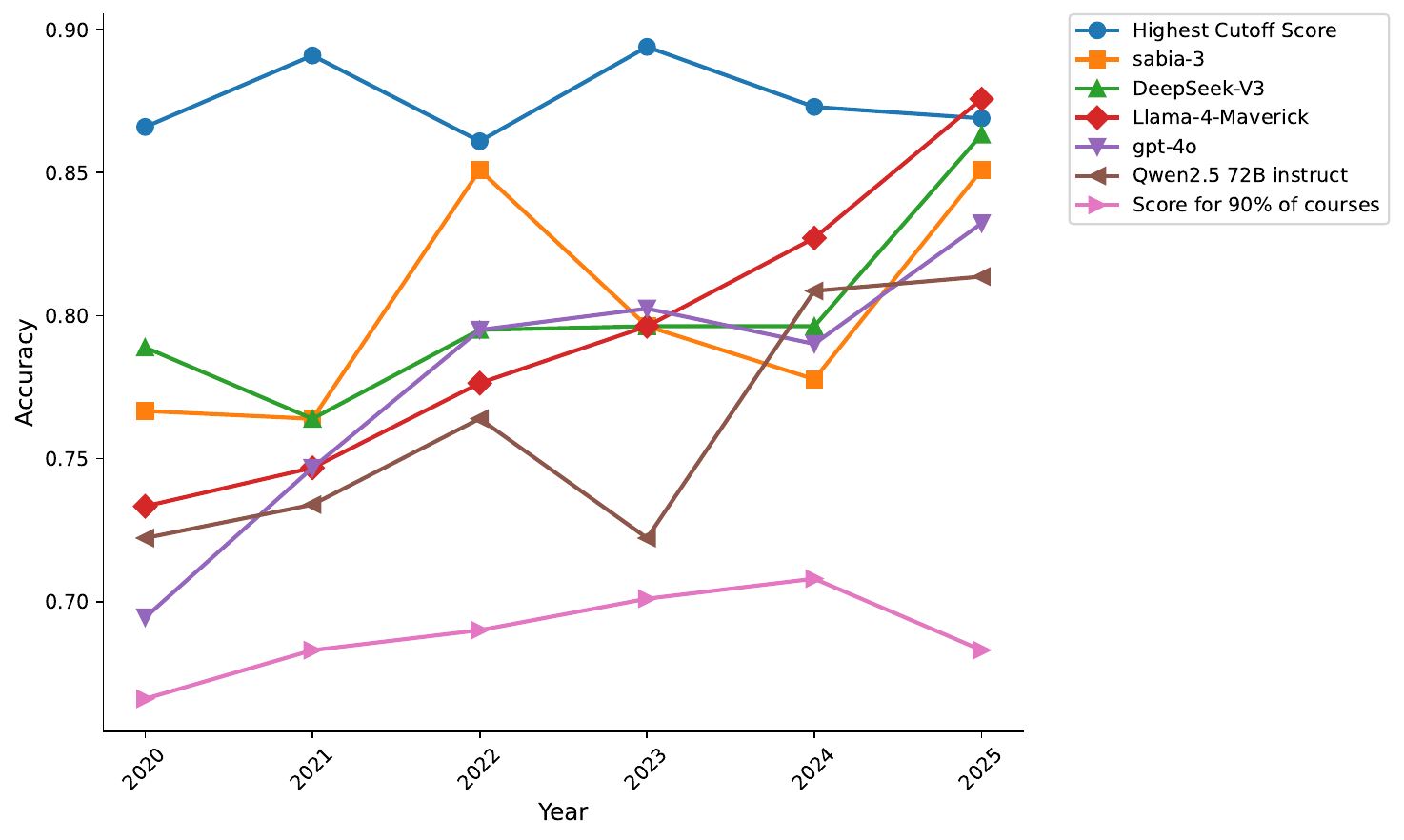}
    \caption{Performance of the top 5 tested models in each year, alongside the Highest Cutoff score and Score for passing in 90\% of the offered courses.}
    \label{fig:perf_over_time}
\end{figure}

All larger models evaluated already achieve scores sufficient to gain admission to approximately 90\% of undergraduate courses at both USP and UNICAMP. Figure~\ref{fig:perf_over_time} illustrates the accuracy over the past six years for the top five models tested when provided with context-aware captions. The pink line represents the threshold score necessary to gain admission to 90\% of courses. The models consistently meet or exceed this bar throughout the observed period, with a slight performance increase noted in recent years. Importantly, considering the exams' recency (particularly the 2025 tests), it is unlikely that most models had encountered these questions during training. This indicates their strong performance stems from genuine reasoning abilities rather than merely memorizing publicly available exam solutions.

However, despite their generally high performance, most models do not achieve scores sufficient to enter the most competitive undergraduate program\footnote{Typically, the most competitive undergraduate program is Medicine.}, represented by the 'Highest Cutoff Score' in the graph. An exception is LLaMA 4 Maverick, which attains a qualifying score in the 2025 exam, though not in previous years.

\section{Conclusion}

In this study, we expanded the BLUEX benchmark dataset by introducing image captions, enabling broader question accessibility for language models without multimodal capabilities. By exploring two captioning strategies--blind and context-aware captions --we found comparable model performance, despite observed differences in caption length. Context captions provided shorter yet more targeted descriptions compared to blind captions, which, although more detailed, contained less selectively relevant information.

Our evaluation across various commercial and open-source language models revealed several insights. Among commercial models, Sabia-3 excelled in text-only scenarios, whereas GPT-4o demonstrated superior performance when handling image-based questions. Within large open-source models, DeepSeek-V3 consistently showed strong results across all evaluated conditions, highlighting the effectiveness of models incorporating large parameter counts or mixture-of-expert architectures. Among smaller models, Qwen 2.5-14B displayed remarkable competitiveness, surpassing its peers significantly.

Importantly, providing captions, especially context-aware ones, notably improved model accuracy on questions that inherently required image interpretation. This performance enhancement was more pronounced in larger models, indicating a correlation between model size and the effective utilization of the textual descriptions.

Our work further develops the Portuguese benchmark landscape by effectively doubling the number of questions usable for non-multimodal LLMs compared to the original benchmark and demonstrates the current capabilities of state-of-the-art models.

Additionally, further research could expand the evaluation scope to include open-ended, dissertative questions, which are also part of the examined exams. This would broaden the assessment of LLM capabilities beyond the current multiple-choice format, providing a richer evaluation framework. Future work could also incorporate exams from other university entrance tests, further diversifying the benchmark.



\bibliographystyle{sbc-template}
\bibliography{sbc-template}

\end{document}